\documentclass[sigconf]{acmart}

\usepackage{booktabs}
\usepackage{enumitem}
\usepackage{xspace}
\usepackage{amstext}
\usepackage{multirow}
\usepackage{array}
\usepackage{subfigure}
\usepackage{colortbl}
\usepackage[ruled,vlined]{algorithm2e}
\usepackage{pifont}    %
\usepackage{amsthm}

\SetKwInput{KwInput}{Input}    %
\SetKwInput{KwRequire}{Require}    %
\SetKwInput{KwOutput}{Output}    %

\definecolor{mygray}{gray}{0.4}

\newcommand{\mytiny}[1]{{\!\footnotesize{\textcolor{mygray}{#1}}}}

\newcommand{\setting}{reference-limited compositional zero-shot learning\xspace}
\newcommand{\Bsetting}{Reference-Limited Compositional Zero-Shot Learning\xspace}
\newcommand{\shortname}{RL-CZSL\xspace}

\newcommand{\datasetattr}{\textsc{\shortname-ATTR}\xspace}
\newcommand{\datasetact}{\textsc{\shortname-ACT}\xspace}

\newcommand{\Ours}{Meta Compositional Graph Learner\xspace}
\newcommand{\shortOurs}{MetaCGL\xspace}

\newcommand{\ie}{\textit{i.e.}}
\newcommand{\eg}{\textit{e.g.}}

\AtBeginDocument{%
  }

\copyrightyear{2023}
\acmYear{2023}
\setcopyright{rightsretained}
\acmConference[ICMR '23]{International Conference on Multimedia
Retrieval}{June 12--15, 2023}{Thessaloniki, Greece}
\acmBooktitle{International Conference on Multimedia Retrieval (ICMR '23), June
12--15, 2023, Thessaloniki, Greece}\acmDOI{10.1145/3591106.3592225}
\acmISBN{979-8-4007-0178-8/23/06}

\begin{document}

\title{Reference-Limited Compositional Zero-Shot Learning}

\author{Siteng Huang}
\affiliation{%
  \institution{Zhejiang University}
}
\email{huangsiteng@westlake.edu.cn}

\author{Qiyao Wei}
\affiliation{%
  \institution{University of Cambridge}
}
\email{qw281@cam.ac.uk}

\author{Donglin Wang}
\authornote{Corresponding author.}
\affiliation{%
  \institution{Westlake University}
}
\email{wangdonglin@westlake.edu.cn}

\begin{abstract}
Compositional zero-shot learning (CZSL) refers to recognizing unseen compositions of known visual primitives, which is an essential ability for artificial intelligence systems to learn and understand the world.
While considerable progress has been made on existing benchmarks, we suspect whether popular CZSL methods can address the challenges of few-shot and few referential compositions, which is common when learning in real-world unseen environments.
To this end, we study the challenging \setting (\shortname) problem in this paper, \ie, given limited seen compositions that contain only a few samples as reference, unseen compositions of observed primitives should be identified.
We propose a novel \Ours (\shortOurs) that can efficiently learn the compositionality from insufficient referential information and generalize to unseen compositions.
Besides, we build a benchmark with two new large-scale datasets that consist of natural images with diverse compositional labels, providing more realistic environments for \shortname.
Extensive experiments in the benchmarks show that our method achieves state-of-the-art performance in recognizing unseen compositions when reference is limited for compositional learning.
\end{abstract}




\maketitle

\section{Introduction}

\setlength{\textfloatsep}{2pt}

\begin{figure}[!t]    %
  \centering
  \includegraphics[width=0.48\textwidth]{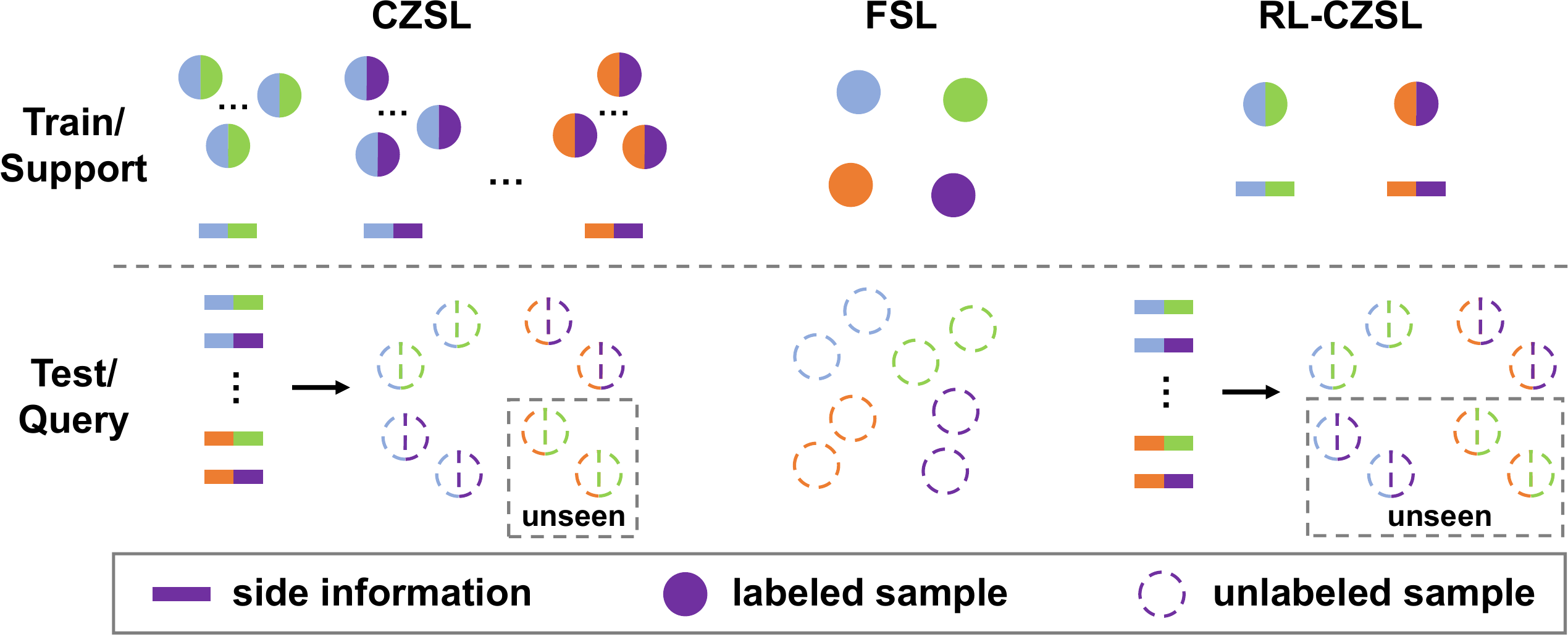}
  \vspace{-6mm}
  \caption{
    Comparison among compositional zero-shot learning (CZSL), few-shot learning (FSL), and our proposed reference-limited compositional zero-shot learning (RL-CZSL). Different colors indicate different categories of entities or primitives.
  }
  \label{fig:setting_comparsion}
\end{figure}

Different from standard systems that are limited to a fixed set of categories at a time, humans generalize to a large, essentially ``unbounded'' concept space by reasoning in a compositional manner~\cite{Bahdanau:systematic-generalization,Vedantam:CURI}.
This method of identifying novel complex concepts by composing known components (which we call the ``\textit{primitives}'' in this paper) is called compositional generalization, representing the essential ability of human intelligence to make ``infinite use of finite means''~\cite{Chomsky:CG-definition,Humboldt:CG-definition}.
For example, based on familiarity with tomatoes and other red objects, people can recognize a red tomato when they first encounter it.
Similarly, it is easy to understand the behavior of cutting a pizza after cutting a cake and knowing what a pizza is.
In the widely studied compositional zero-shot learning (CZSL) problem, with the side information (\textit{e.g.}, attributes, textual descriptions, and label embeddings), visual perception models are also expected to identify unseen compositional concepts, that is, they learn the compositionality of primitives from sufficient training samples and are tasked with generalizing to unseen combinations of these primitives~\cite{Misra:red-wine-to-red-tomato,Purushwalkam:task-driven-modular-networks,Naeem:CGQA}.

While these efforts have contributed to a more comprehensive perception of the world, we argue that the existing setup seems idealistic and inappropriate to simulate natural human learning, and two core challenges should be considered when evaluating compositional learners.
(1) \textbf{Few-shot}: Humans have an inherent ability to learn the compositionality of complex concepts with only a few examples and transfer %
the learned knowledge to different situations. %
However, AI systems will suffer from severer generalization issues if training samples are insufficient, as the empirical risk is far from being a good approximation for expected risk~\cite{Wang:FSL-survey}.
Although an increasing number of models have tried to alleviate potential overfitting~\cite{Snell:ProtoNet,Finn:MAML,Chen:CloserLookFSC}, they still treat every class as an independent entity and require referential data for any novel concept.
Hence we would like to investigate whether compositional learning can be performed with restricted sample size, in other words, whether few-shot learners can generalize to unseen label compositions.
(2) \textbf{Few referential compositions}: 
Unlike recent methods~\cite{Saini:OADis,Tian:IVR,Li:SCEN} that have to refer to multiple combinations with the same primitive to extract semantic invariants from them,
humans can discover potential primitives from a few combinations, or even only one, based on prior knowledge. %
This contributes to the adaptation of humans to the long-tailed distribution of various compositional concepts in the real world, \ie, there exist a few common primitives and many more composition-scarce primitives, making collecting all possible scenarios for each primitive in advance expensive and time-consuming.
Therefore, 
few referential compositions should also be a natural constraint for human-level compositional learning.

In this paper, we introduce an untouched problem, \textbf{\setting (\shortname)}, to 
approximate real-world situations that would be encountered when 
compositional learning is required.
The term ``reference-limited'' is adopted to indicate that when the model performs compositional learning, the combinations that can be used as references are limited in terms of the number of both categories and labeled samples, as shown in \figurename~\ref{fig:setting_comparsion}.
Therefore, \shortname requires the learner to incorporate appropriate priors into learning,
so that it can quickly learn the compositionality clues without superfluous references.
Furthermore, due to the lack of rich categories and types of primitives and compositions, the existing datasets
can not afford to create a large amount of testing environments for a comprehensive measurement.
To address the limitations of the datasets and provide suitable conditions for systematic comparisons on \shortname task, we build two benchmark datasets that consist of over 99k and 30k natural images covering sufficient attribute-object and action-object compositional labels, supporting us to sample realistic episodes to simulate partially observable worlds.

To address the new challenges, we propose a novel method \textbf{\Ours (\shortOurs)}.
\shortOurs constructs a compositional graph to learn the dependencies between primitive and composition representations, and learns better semantic embeddings by aggregating information of neighbor nodes.
With the updated semantic embeddings, \shortOurs generates a prior correlation map to estimate which features are related to the prediction target.
Moreover, \shortOurs applies an effective bi-level optimization strategy during training, which contains a simple data augmentation method named Compositional Mixup to enhance the generalization ability.    %
We compare our \shortOurs with representative CZSL methods on the proposed datasets, and the results show that \shortOurs significantly outperforms prior methods in recognizing unseen compositions.
By shedding light on the limitations of existing settings and approaches, we hope to spur future work to develop human-level compositional generalization ability for intelligent systems.
In summary, our contributions are as follows:

1. We introduce a new problem named \setting (\shortname), where given only a few samples of limited compositions, the model is required to generalize to recognize unseen compositions. This offers a more realistic and challenging environment for evaluating compositional learners.

2. We establish two benchmark datasets with diverse compositional labels and well-designed data splits, providing the required platform for systematically assessing progress on the task.

3. We propose a novel method, \Ours (\shortOurs), for the challenging \shortname problem. Experimental results show that \shortOurs consistently outperforms popular baselines on recognizing unseen compositions.
\vspace{-2mm}

\section{Related Work} \label{related_work}

\textbf{Compositional Zero-shot Learning (CZSL)} aims to recognize unseen attribute-object compositions at test time while each constituent exists in training samples, given side information that describes novel composition pairs, \textit{e.g.}, word embeddings, attribute annotations, or text descriptions.
As the early Visual Product (VisProd)~\cite{Misra:red-wine-to-red-tomato} baseline just computes the outputs of individual primitive classifiers as the predicted probability of the corresponding composition, some notable works~\cite{Misra:red-wine-to-red-tomato,Li:symmetry-and-group,Mancini:CompCos,Naeem:CGQA} argue that compositionality requires learning a joint compatibility function between the image, the attribute, and the object.
And recently, state-of-the-art methods~\cite{Saini:OADis,Tian:IVR,Li:SCEN} rely on comparing different compositions of the same primitive to learn to disentangle visual features.
However, all of these works assume that there exist sufficient referential compositions and samples for learning in a compositional manner.
In this paper, we propose the \setting (\shortname) problem that removes this assumption to be more close to the real-world unseen environments.
And our experimental results show that the state-of-the-art CZSL methods also struggle with this new challenge.
As our introduced \shortname benchmark datasets also contain compositional labels of action-object pairs, we also compare our topic with human-object interaction (HOI) detection~\cite{Kato:HOI,Hou:HOI-detection}, which aims at detecting all the human-interaction-object triplets in an image.
This task can also be viewed as a compositional zero-shot learning problem, as test images may contain interaction-object pairs that do not appear in the training data.
However, HOI detection methods rely on a pre-trained object detector to localize the human and objects for further processing,
which is not available for novel primitives in unseen environments.

\noindent \textbf{Few-shot Learning (FSL)} requires learning new tasks with few labeled examples.
Recent FSL advances can be roughly categorized into the following three groups:
(1) \textit{metric-based} methods \cite{Vinyals:MatchNet,Snell:ProtoNet,Sung:RelationNet} learn a generalizable embedding model to transform all samples into a common metric space, where simple classifiers can be executed directly.
(2) \textit{initialization-based} methods~\cite{Finn:MAML,Raghu:ANIL} learn a good set of initial parameters for the whole model or part of it, so that the model can quickly adapt to novel classes in a small number of gradient update steps.
(3) \textit{pretraining-based} methods~\cite{Chen:CloserLookFSC,Dhillon:FSL-Baseline,Tian:FSL-Baseline} train a feature extractor with all the training data, and fix it during the meta-test phase whilst learning new classifiers for novel classes.
Recently, several FSL works~\cite{Tokmakov:Comp-Feats,Zou:CFSL,Huang:AGAM}     %
have aimed to improve the generalization performance with compositional representations.
Limited by the traditional FSL setting on which they are based, these methods only consider feature compositionality and have not explored how to generalize to new label compositions. %

\vspace{-2mm}
\section{\Bsetting}

\subsection{Problem Formulation}

The ultimate goal of the \shortname task is to recognize unseen visual pair compositions, whose primitives have only appeared in limited seen compositions containing only a few samples.
In this paper, we follow the FSL setting to use the sampled episodes as a simulation of independent test environments,
which refer to the data for learning as \textit{support} and the data for inference as \textit{query}.
In addition, we apply an \textbf{open world} setting that while all compositions contained in the \textit{support classes} are \textit{seen} ones, the \textit{query classes} include not only \textit{unseen compositions}, but also \textit{seen compositions}.
At the same time, no constraint on the test time search space is imposed.
Allowing predictions to come from all possible pairs in the current episode, the setting is more close to the unseen environments that are likely to arise in real-world deployments, and thus leads to a more comprehensive study on achieving a balanced and promising performance of both seen and unseen compositions.

More formally, we consider the visual recognition setting where each image $x$ is associated with a complex concept $c$ that is a pair composition of two primitives $p^1$ and $p^2$, \textit{i.e.}, $c = (p^1, p^2)$. For example, $p^1$ can represent a state like ``cooked'' or an action like ``cut'', while $p^2$ can refer to an object such as ``chicken'' or ``pizza''.
When testing, the model are evaluated on episodes that are sampled from a set of \textit{novel} data $\mathcal{D}_n = \{(x_n^{(i)}, c_n^{(i)})\}$ with label space $\mathcal{C}_n \subset \mathcal{P}_n^1 \times \mathcal{P}_n^2 = \{(p_n^1, p_n^2) | p_n^1 \in \mathcal{P}_n^1, p_n^2 \in \mathcal{P}_n^2\}$.
$\mathcal{C}_n$ denotes the novel composition set, $\mathcal{P}_n^1$, $\mathcal{P}_n^2$ are the two corresponding novel primitive sets with different primitive types (\ie, attributes, actions, or objects), and each primitive set contains $N^p$ primitive categories.
Each episode contains a \textit{support} set $\mathcal{S} = \{(x_s^{(i)}, c_s^{(i)}) | i = 1, 2, \dots, N^c_s \times K^c_s\}$ that consists of $N^c_s$ support classes with $K^c_s$ labeled samples per class, and a \textit{query} set $\mathcal{Q} = \{(x_q^{(i)}, c_q^{(i)}) | i = 1, 2, \dots, N^c_q \times K^c_q\}$ that consists of $N^c_q$ query classes with $K^c_q$ samples per class.
The query classes not only contain $N^c_s$ \textit{seen} compositions that are all in the support classes, but also comprise $(N^c_q - N^c_s)$ \textit{unseen} compositions that do not overlap with seen compositions.
However, seen and unseen compositions in the same episode share the same two primitive sets sampled from $\mathcal{P}_n^1$ and $\mathcal{P}_n^2$, providing the possibility for unseen compositions to be recognized.  %
Following the open world setting, the prediction space of the model contains ${N^p} \times {N^p}$ compositions including seen, unseen and unfeasible ones.
And the goal of the model is to correctly predict the compositional labels of samples in $\mathcal{Q}$ with the access to $\mathcal{S}$.

To extract the prior knowledge for learning to rapidly separate primitive features from images,
in the training phase, the model possesses the access to a set of \textit{base} data $\mathcal{D}_b = \{(x_b^{(i)}, c_b^{(i)})\}$ with label space $\mathcal{C}_b \subset \mathcal{P}_b^1 \times \mathcal{P}_b^2$.
Note that the base and novel primitive sets do not overlap, \textit{i.e.}, $\mathcal{P}_b^1 \cap \mathcal{P}_n^1 = \emptyset$ and $\mathcal{P}_b^2 \cap \mathcal{P}_n^2 = \emptyset$, and thus $\mathcal{C}_b \cap \mathcal{C}_n = \emptyset$ also holds.
We would like to mention that \shortname does not require a specific procedure for learning from the training data, that is, episodic and non-episodic methods are both allowed.
\vspace{-2mm}

\subsection{Proposed Benchmarks}

\begin{table}[!t]    %
  \centering
  \caption{Basic statistics of our proposed datasets. The symbol \# is used as an abbreviation for ``number of''. }
  \vspace{-2mm}
  \resizebox{\linewidth}{!}{
    \begin{tabular}{ccc}
    \toprule
    \textbf{Dataset} & \textbf{\shortname-ATTR} & \textbf{\shortname-ACT} \\
    \midrule
    \rowcolor[rgb]{ .949,  .949,  .949} \textbf{Composition type} & attribute-object & action-object \\
    \textbf{Total \#$c$} & 1,768 & 574 \\
    \rowcolor[rgb]{ .949,  .949,  .949} \textbf{\#$c$ in train / val / test} & 1,076 / 136 / 556 & 214 / 22 / 338  \\
    \textbf{Total \#$p^1$} & 190   & 185 \\
    \rowcolor[rgb]{ .949,  .949,  .949} \textbf{\#$p^1$ in train / val / test} & 105 / 33 / 52 & 52 / 10 / 123 \\
    \textbf{Total \#$p^2$} & 488   & 154 \\
    \rowcolor[rgb]{ .949,  .949,  .949} \textbf{\#$p^2$ in train / val / test} & 281 / 12 / 195 & 59 / 11 / 84 \\
    \textbf{Total \#samples} & 99,771 & 30,420 \\
    \rowcolor[rgb]{ .949,  .949,  .949} \textbf{\#samples in train / val / test} & 51,928 / 29,922 / 17,921 & 20,604 / 1,207 / 8,609 \\
    \bottomrule
    \end{tabular}%
  }
  \label{tab:dataset_stats}%
\end{table}%

Although several CZSL datasets like MIT-States~\cite{Isola:MIT}, UT-Zap50K~\cite{Yu:UTZap}, and C-GQA~\cite{Naeem:CGQA}, have been proposed, there exist limitations that prevent them from becoming appropriate benchmarks for measuring human-level compositional generalization ability: 
(1) They contain images with only state-object pair labels, failing to cover the most frequent types of compositions in the real world.
We argue that another common composition type, action-object pairs, should also be considered to examine the compositional reasoning ability of models. 
(2) More importantly, we found that when trying to divide them into three splits, existing datasets did not have enough categories of compositions and primitives to ensure that no primitives in the splits of different phases overlapped. 
This made it impossible to create a sufficient number of various episodes to simulate diverse unseen environments during the test phase, ultimately reducing the validity of the evaluation.

\begin{figure}[!t]   %
  \centering 
  \subfigure[\datasetattr: train]{\label{subfig:C-GQA}{
  \includegraphics[width=0.28\linewidth]{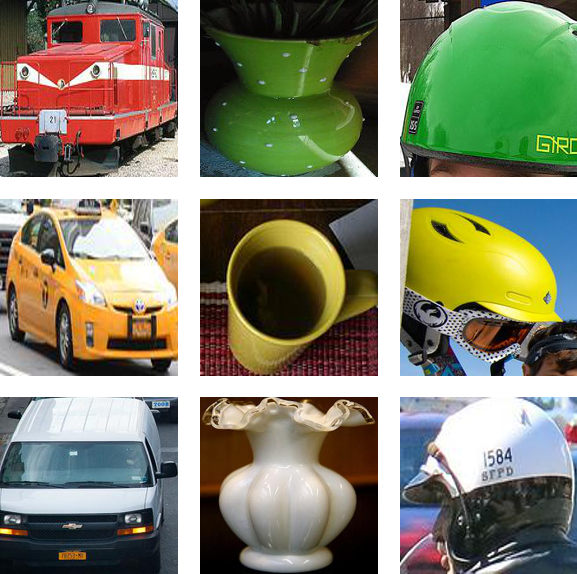} 
  } }
  \subfigure[\datasetattr: val]{\label{subfig:UT-Zappos50K}{
  \includegraphics[width=0.28\linewidth]{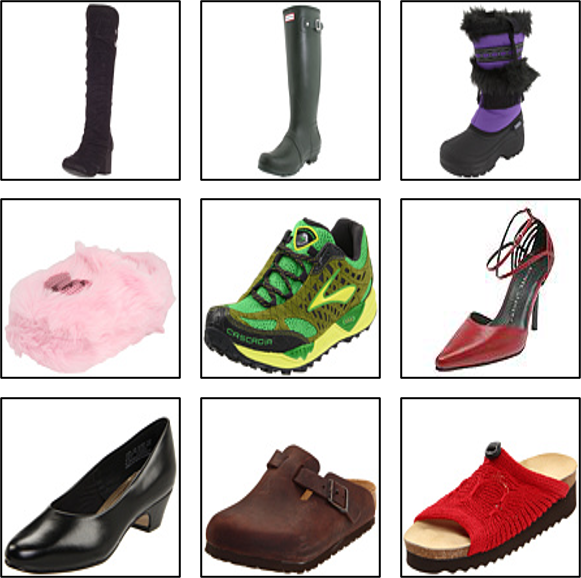} 
  } }
  \subfigure[\datasetattr: test]{\label{subfig:MIT-States}{
  \includegraphics[width=0.28\linewidth]{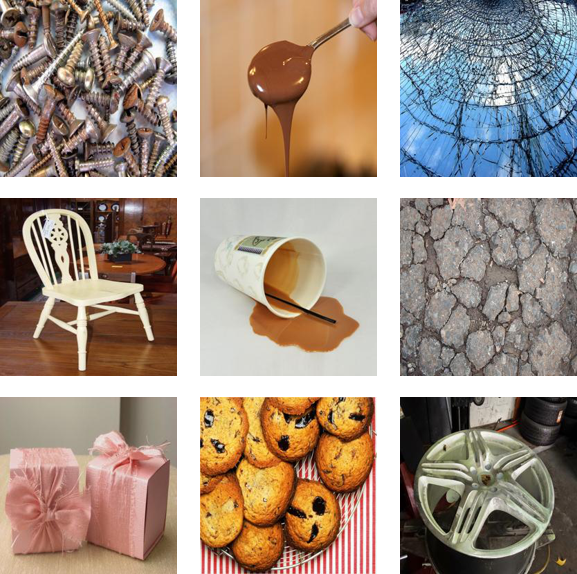} 
  } }
  \subfigure[\datasetact: train]{\label{subfig:HICO}{
  \includegraphics[width=0.28\linewidth]{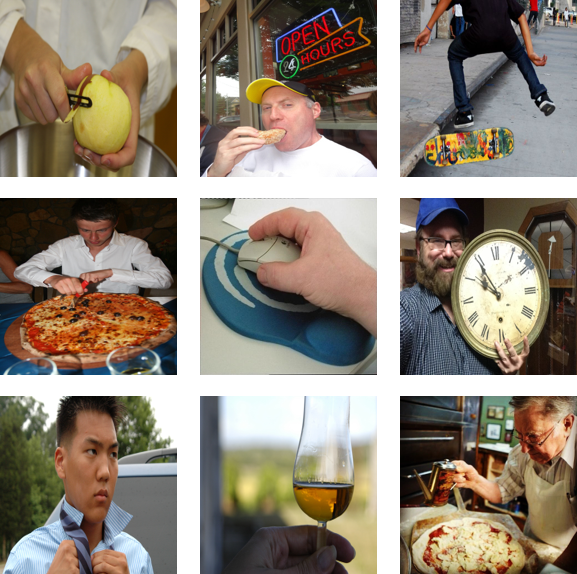} 
  } }
  \subfigure[\datasetact: val]{\label{subfig:Visual_Genome}{
  \includegraphics[width=0.28\linewidth]{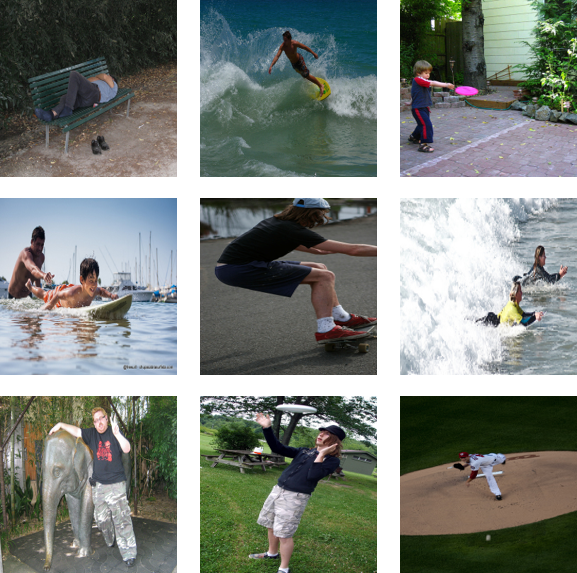} 
  } }\
  \subfigure[\datasetact: test]{\label{subfig:imSitu}{
  \includegraphics[width=0.28\linewidth]{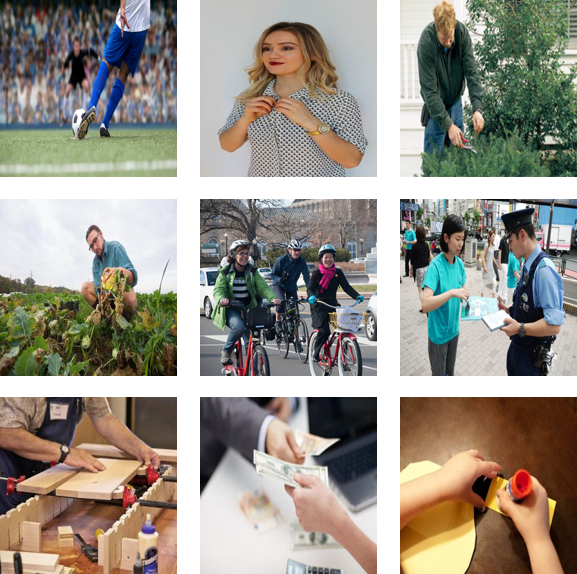} 
  } }
  \vspace{-4mm}
  \caption{Examples taken from \datasetattr (the top row) and \datasetact (the bottom row) benchmark datasets.}
  \label{fig:dataset_examples}
\end{figure}

Therefore, we refine public datasets to form two suitable benchmark datasets for \shortname, named \textbf{\datasetattr} and \textbf{\datasetact}.
\datasetattr contains over 99k images attached with 1,768 attribute-object pair labels from C-GQA~\cite{Naeem:CGQA}, UT-Zap50K~\cite{Yu:UTZap}, and MIT-States~\cite{Isola:MIT}.
And \datasetact consists of over 30k images with 574 action-object pair labels from HICO~\cite{Chao:HICO}, Visual Genome~\cite{Krishna:VisualGenome}, and imSitu~\cite{Yatskar:imSitu}.
The statistics are summarized in \tablename~\ref{tab:dataset_stats}, and we show some sample images in \figurename~\ref{fig:dataset_examples}.
\vspace{-2mm}

\paragraph{Organizing Data}
To ensure that the two datasets we obtained meet the needs of the \shortname task, we filtered the data in all splits from the perspective of labels.
Specifically, (1) compositions with fewer than 10 samples were screened out to ensure that enough same-class support and query samples can be simultaneously sampled without duplicates,
(2) for primitives that appeared in multiple splits, we kept them in at most one split,
(3) size-related attribute primitives that cannot be accurately depicted in the images, such as ``small'', ``large'', ``long'' and ``short'', were also filtered out from our datasets.
Specially, images from the Visual Genome dataset are densely annotated with numerous attributes and objects, lacking a description of the focus of the content.
Therefore, we kept the attribute-object compositional label of the largest bounding box in each image, which most likely corresponds to the main content, and removed other annotations.
\vspace{-2mm}

\paragraph{Episode Sampling Strategy}   \label{episode_sampling_strategy} 

In this section, we outline the sampling strategy that creates more realistic episodes for the \shortname task, and the corresponding pseudocode is illustrated in Algorithm~\ref{algo:episode_sampling}.
In each experiment, the value of $N^p$ is fixed for all sampled episodes in the same phase.
However, we allow episodes to have a different number of seen and unseen compositions, \textit{i.e.}, the values of $N^c_s$ and $N^c_q$ may vary from episode to episode.
As $N^c_q$ actually corresponds to the number of all potential compositions that can be obtained by pairing all primitives in the episode and also exist in the dataset, 
$N^c_s$ is randomly sampled within a certain range of [$N^p+1$, $N^c_q-1$], making the episode closer to reality.
The maximum value of this range guarantees that there exist unseen compositions in the episode, and the minimum value implies that each primitive has the opportunity to appear in more than one seen composition without being bound to another primitive all the time.

Concretely, for each episode, we first randomly sample $N^p$ seen compositions to obtain two primitive sets $\mathcal{P}_n^1$, $\mathcal{P}_n^2$ without duplicate primitives.
Then, we check if $\mathcal{P}_n^1$ and $\mathcal{P}_n^2$ can be paired to get enough existing compositions for being divided into seen and unseen ones.
An episode that can achieve the required number of composition pairs will be regarded as a valid episode, and the remaining compositions will be randomly assigned to seen and unseen groups on the premise of satisfying the restriction of $N^c_s$.
Therefore, we have $N^c_s$ seen compositions for the support classes and $N^c_q$ compositions including seen and unseen ones for the query classes. 
Next, we randomly sample $K^c_s$ support samples for each seen composition and $K^c_q$ samples for each composition in the query classes.
Note that for seen compositions that exist in both support and query classes, the samples assigned to the two sets are not duplicated.
\vspace{-2mm}

\begin{algorithm}[!t]    %
    \caption{Episode Sampling in \shortname}
    \label{algo:episode_sampling}
    \DontPrintSemicolon
    \KwRequire{$\mathcal{D}$ with label space $\mathcal{C}$ according to the requested class split of the given dataset, $N^p$, $K^c_s$, $K^c_q$}
    \KwOutput{the sampled episode $( \mathcal{S}, \mathcal{Q} )$}
    
    \textcolor{blue}{\textbf{Step 1.} Sample primitive sets $\mathcal{P}^1$, $\mathcal{P}^2$}
    
    $\mathcal{P}^1$ = \{\}, $\mathcal{P}^2$ = \{\}, seen compositions $\mathcal{C_\text{seen}}$ = \{\}\;
  
    \While{$|\mathcal{C_\text{seen}}| < N^p$}
      {
        Randomly sample a composition $c = (p^1, p^2)$ from $\mathcal{C}$\;
        \If{($p^1$ not in $\mathcal{P}^1$) and ($p^2$ not in $\mathcal{P}^2$)}{
          Add $c$ into $\mathcal{C_\text{seen}}$, $p^1$ into $\mathcal{P}^1$, and $p^2$ into $\mathcal{P}^2$\;
          }
      }
    
    \textcolor{blue}{\textbf{Step 2.} Sample the support set $\mathcal{S}$}
    
    $\mathcal{S}$ = \{\}, candidate compositions $\mathcal{C_\text{candidate}}$ = \{\}, unseen compositions $\mathcal{C_\text{unseen}}$ = \{\}\;
    \For{$c \in \mathcal{P}^1 \times \mathcal{P}^2$}{
      \If{($c$ in $\mathcal{C}$) and ($c$ not in $\mathcal{C_\text{seen}}$)}{
        Add $c$ into $\mathcal{C_\text{candidate}}$\;
      }
    }
  
    \If(\tcp*[f]{Seen and unseen compositions are insufficient}){$|\mathcal{C_\text{candidate}}| < 2$}{
      Jump back to \textbf{Step 1}\;
    }

    Randomly assign the first two compositions in $\mathcal{C_\text{candidate}}$ to each of $\mathcal{C_\text{seen}}$ and $\mathcal{C_\text{unseen}}$, and the remaining ones in $\mathcal{C_\text{candidate}}$ are randomly assigned to either $\mathcal{C_\text{seen}}$ or $\mathcal{C_\text{unseen}}$ each time\;
  
    \For{$c \in C_\text{seen}$}{
      Randomly sample $K^c_s$ samples of $c$ into $\mathcal{S}$\;
    }

    \textcolor{blue}{\textbf{Step 3.} Sample the query set $\mathcal{Q}$}
  
    $\mathcal{Q}$ = \{\}\;
  
    \For{$c \in C_\text{seen}$}{
      Randomly sample $K^c_q$ samples of $c$ from those do not overlap with $\mathcal{S}$ into $\mathcal{Q}$\;
    }
  
    \For{$c \in C_\text{unseen}$}{
      Randomly sample $K^c_q$ samples of $c$ into $\mathcal{Q}$\;
    }
\end{algorithm}

\paragraph{Evaluation Metrics}
We use three metrics to evaluate how well the learned model recognizes both unseen and seen composition pairs, consistent with the adopted open world setting:
(1) \textbf{Unseen accuracy (UA)}: The average of the accuracy computed on query samples from unseen compositions on all test episodes.
(2) \textbf{Seen accuracy (SA)}: The average of the accuracy computed on query samples from seen compositions on all test episodes.
(3) \textbf{Harmonic mean (HM)}: A metric that quantifies the overall performance of both seen and unseen accuracy based on the results of all test episodes, defined as: $\text{HM}=2\left( \text{SA} *  \text{UA}\right) /\left( \text{SA}+ \text{UA}\right)$. 
\vspace{-2mm}

\section{\Ours}

\begin{figure}[!t]    %
  \centering
  \includegraphics[width=0.48\textwidth]{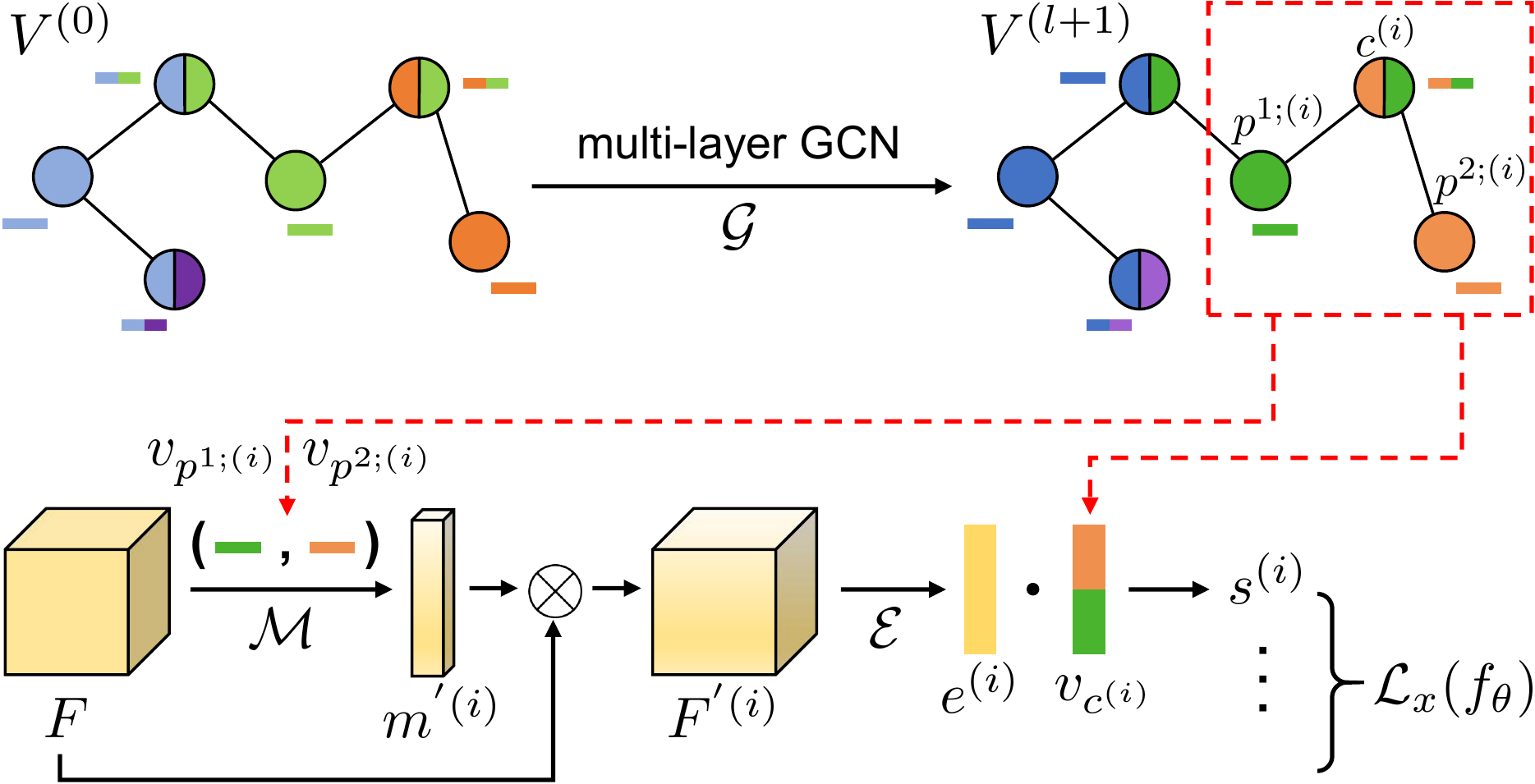}
  \vspace{-6mm}
  \caption{
    The architecture of our proposed \Ours (\shortOurs). Best viewed in color.
  }
  \label{fig:method}
  \vspace{-3mm}
\end{figure}

For \shortname, we propose a novel \Ours (\shortOurs), whose architecture is illustrated in \figurename~\ref{fig:method}.
Following the popular CZSL framework~\cite{Mancini:CompCos,Naeem:CGQA}, \shortOurs 
embeds both images and composition representations into a shared embedding space and computes a compatibility score to measure the similarity between visual and composition embeddings.
Also, \shortOurs follows the episodic training paradigm to keep the spirit of matching training and test conditions, \ie, \shortOurs is trained on episodes sampled from $\mathcal{D}_b$ using the same algorithm as used for test episodes.
In the following, we introduce how \shortOurs learns composition and visual embeddings, calculates compatibility scores, and is trained with a designed optimization process.

\vspace{-2mm}
\subsection{Learning Composition Embeddings}

We choose to learn semantic embeddings for compositions and primitives with a pre-defined compositional graph, as the graph structure has been proved to be effective in modeling dependency relationships between them~\cite{Naeem:CGQA}.
Specifically, we construct a graph that contains $H = |\mathcal{P}^1| + |\mathcal{P}^2| + |\mathcal{C}|$ nodes in each episode, where $\mathcal{P}^1$ and $\mathcal{P}^2$ correspond to the primitive sets in the episode, and $\mathcal{C}$ is the set of all potential pairs composed of primitives in $\mathcal{P}^1$ and $\mathcal{P}^2$, including existent and nonexistent compositions in the dataset.
The features of primitive nodes are initialized with the pre-trained word embeddings to utilize the prior knowledge extracted from large corpora, and the features of composition nodes are initialized by averaging the word embeddings of associated primitives. 
We represent the initial features of all nodes as $V^{(0)} \in \mathbb{R}^{H \times d_w}$, where $d_w$ denotes the dimension of the word embeddings.
The nodes of $p^1$, $p^2$ and $c$ are connected one by one for each $c = (p^1, p^2) \in \mathcal{C}$, 
and a self-loop is added to each node.
Thus, a symmetric adjacency matrix $A \in \mathbb{R}^{H \times H}$ can be obtained by setting $A_{ij} = 1$ if nodes $i$ and $j$ are connected, otherwise $A_{ij} = 0$.
By applying a multi-layer graph convolutional network~\cite{Kipf:GCN} $\mathcal{G}$, the node features can be updated as 

\begin{align}
V^{(l+1)} = \sigma (D^{-\frac{1}{2}}AD^{-\frac{1}{2}}V^{(l)}\theta_\mathcal{G}).
\end{align}

\noindent Here the non-linear activation function $\sigma$ is ReLU, $V^{(l+1)}$ is the output of the $l^{th}$ layer, $\theta_\mathcal{G}$ %
is the trainable weight matrix.
$D \in \mathbb{R}^{H \times H}$ is a diagonal node degree matrix with $D_{ii} = \sum_j A_{ij}$, which preserves the scale of feature vectors to avoid gradient vanishing or explosion.    %
\vspace{-2mm}

\subsection{Learning Visual Embeddings}

To map the image features $F \in \mathbb{R}^{c \times h \times w}$ extracted by the backbone to the shared embedding space, an additional embedding function is required.    %
While prior methods~\cite{Mancini:CompCos,Naeem:CGQA} use simple multi-layer perceptrons (MLPs) to directly embed image features into a vector, we argue that features semantically unrelated to the target composition, such as the background, are also encoded into the embedding.
To avoid such noise affecting the learning of the shared embedding space and thus damaging the accuracy of recognition, we propose to make a prior prediction about the features that might be relevant to the recognition target.
Specifically, we transform the learned semantic embeddings into a prior correlation map with a correlation map generating network $\mathcal{M}$. 
Such a correlation map tells the probability of the features belonging to the target composition.
For the $i$-th composition $c^{(i)} = (p^{1;(i)}, p^{2;(i)})$, we obtain the corresponding composition and primitive embeddings $v_{c^{(i)}}, v_{p^{1;(i)}}, v_{p^{2;(i)}} \in \mathbb{R}^{d}$ from the outputs of $\mathcal{G}$, where $d$ is the dimension of the output vectors.
In each episode, more images contain the same primitive than those that are associated with the same composition, \ie, more samples can be used to learn the visual invariants for the primitive embeddings.
Therefore, we leverage primitive embeddings instead of composition embeddings to produce the prior correlation map.
For each primitive embedding $v_{p}$, we firstly apply global average pooling (GAP) on the feature map $F$, and concatenate $\text{GAP}(F)$ with $v_{p}$ to obtain a new vector, which is then forwarded to $\mathcal{M}$.
In practice, $\mathcal{M}$ consists of two convolutions with kernel size 1 and can also be viewed as two linear transformations with a ReLU activation in between.
By exploiting the relationships between each dimension of the semantic vector and each channel of visual vector, $\mathcal{M}$ generates a correlation map $m \in \mathbb{R}^{c}$.
We mention that the parameters $\theta_\mathcal{M}$ are not shared between different primitive types, and we apply an element-wise summation to the outputs of different $\mathcal{M}$ to produce the final correlation map $m^{'} \in \mathbb{R}^{c}$, which is used to highlight the relevant channel.
In summary, we have

{\setlength\abovedisplayskip{-2mm}
\setlength\belowdisplayskip{2mm}
\begin{align}
  m^{'(i)} = \sigma&(\mathcal{M}^1([\text{GAP}(F);v_{p^{1;(i)}}]) \\ + &\mathcal{M}^2([\text{GAP}(F);v_{p^{2;(i)}}])),
\end{align}
}

\noindent where the activation function $\sigma$ is sigmoid, and $[;]$ denotes the operation of concatenating.
Therefore, we have $F^{'(i)} = m^{'(i)} \otimes F$, where channel-wise correlation map values are broadcasted along the spatial dimension, and $\otimes$ denotes element-wise multiplication.
The reason we focus on the channel dimension of the feature map rather than the spatial dimension is that,
the semantics of different primitives are inherently entangled in the spatial dimension,
\ie, it would be difficult to determine which primitive is represented by a single pixel.
However, as each channel of a high-level feature map can be considered as a feature detector~\cite{Zeiler:channel-as-detector}, the correlations between channels and primitive semantics can be more easily learned.
Another embedding function $\mathcal{E}$, implemented by a MLP with parameters $\theta_\mathcal{E}$, is then used to encoder $F^{'(i)}$ into a vector $e^{(i)}$.
\vspace{-2mm}

\subsection{Calculating Compatibility Score}

For each $c^{(i)} \in \mathcal{C}$ in the current episode, we now can calculate a compatibility score $s^{(i)} = e^{(i)} \odot v_{c^{(i)}}$, where $\odot$ represents element-wise dot product.
Naturally, if $c^{(i)}$ is the ground-truth compositional label of the current image $x$, $s^{(i)}$ is expected to be larger and can be optimized with a cross-entropy loss function:
\begin{align}
\label{eq4}  \mathcal{L}_x(f_\theta) = -\log\left(\frac{\exp(s^{(i)})}{\sum_j^{|\mathcal{C}|} \exp(s^{(j)}) } \right),
\end{align}
where $\theta=\{\theta_\mathcal{G}, \theta_\mathcal{M}, \theta_\mathcal{E}\}$ represents all the trainable parameters in \shortOurs, and $f_\theta$ represents the whole model with parameters $\theta$.
And during inference, the prediction can be made by applying a $\text{argmax}$ operation on all compatibility scores calculated for $x$.

\subsection{Training Strategy}

\begin{algorithm}[!t]    %
    \caption{Training and Inference of \shortOurs}
    \label{algo:training_inference}
    \DontPrintSemicolon
    
    \textcolor{blue}{\textbf{Training}}
    
    \KwRequire{base data $\mathcal{D}_b$}
    \KwOutput{trained model with parameters $\theta$}
    Randomly initialize $\theta$\;

    \While{not done}
      {
        Randomly sample a \textbf{train} episode $( \mathcal{S}, \mathcal{Q} )$ from $\mathcal{D}_b$\;
        Calculate $\mathcal{L}_{\mathcal{S}}(f_\theta)$ using Eq.~\eqref{eq4}\;
        Calculate updated parameters $\theta^{'}$ using Eq.~\eqref{eq7}\;
        Generate $\tilde{\mathcal{Q}}$ using Eq.~\eqref{eq5} and Eq.~\eqref{eq6}\;
        Calculate $\mathcal{L}_{\mathcal{\tilde{\mathcal{Q}}}}(f_{\theta^{'}})$ using Eq.~\eqref{eq4}\;
        Update $\theta$ using Eq.~\eqref{eq8}\;
      }
  
    \textcolor{blue}{\textbf{Inference}}
    
    \KwRequire{trained model with parameters $\theta$}
    Calculate $\mathcal{L}_{\mathcal{S}}(f_\theta)$ using Eq.~\eqref{eq4} with $\mathcal{S}$ in the \textbf{test} episode $( \mathcal{S}, \mathcal{Q} )$\;
    Calculate updated parameters $\theta^{'}$ using Eq.~\eqref{eq7}\;
    Use $\theta^{'}$ for the inference of samples in $\mathcal{Q}$
\end{algorithm}

As \shortname requires the model to generalize to unseen compositions with insufficient samples from only a few referential compositions, the training process also requires a special design to avoid overfitting.
We apply a bi-level optimization for \shortOurs, as such an optimization strategy has been proven to be effective in helping to quickly adapt to new environments on a range of problems~\cite{Finn:MAML,Li:MAML-DG}.
Moreover, we propose a new data augmentation method Compositional Mixup to enhance the generalization and robustness.
Based on the popular data augmentation method Mixup~\cite{Zhang:MixUp}, Compositional Mixup improves the process of generating labels to satisfy the needs of compositional learning.

Specifically, in a train episode $(\mathcal{S}, \mathcal{Q})$ sampled from base data $\mathcal{D}_b$, we leverage $\mathcal{Q}$ to construct a set of augmented query samples $\tilde{\mathcal{Q}} = \{(\tilde{x}_q^{(i)}, \tilde{c}_q^{(i)}) | i = 1, 2, \dots, N^c_q \times K^c_q\}$.
For each query sample $(x_q^{(i)}, c_q^{(i)})$ in $\mathcal{Q}$, we randomly sample another query sample $(x_q^{(j)}, c_q^{(j)})$ from $\mathcal{Q}$. 
A new example of image can be formed by a weighted linear interpolation of $x_q^{(i)}$ and $x_q^{(j)}$:
\begin{align}
\label{eq5}  \tilde{x}_q^{(i)} = \lambda x_q^{(i)} + (1 - \lambda) x_q^{(j)},
\end{align}
where $\lambda \in [0, 1]$ is a random value drawn from $\text{Beta}(\alpha, \alpha)$ distribution, and the hyper-parameter $\alpha$ is set to 1.0 in our experiments.
Specially, considering that $c_q^{(i)} = (p^{1;(i)}, p^{2;(i)})$ and $c_q^{(j)} = (p^{1;(j)}, p^{2;(j)})$ are compositional labels, we form the new augmented label as
\begin{align}
\label{eq6}  \tilde{c}_q^{(i)} = \lambda^2 c_q^{(i)} &+ \lambda (1 - \lambda) c_q^{(ij)} \notag \\ 
    &+ \lambda (1 - \lambda) c_q^{(ji)} + (1 - \lambda)^2 c_q^{(j)},
\end{align}
where $c_q^{(ij)} = (p^{1;(i)}, p^{2;(j)})$, $c_q^{(ji)} = (p^{1;(j)}, p^{2;(i)})$.
An intuitive explanation of our Compositional Mixup is that it implicitly introduces new compositions $c_q^{(ij)}$ and $c_q^{(ji)}$ that may not appear in the episode or even in the dataset.
And the bi-level optimization process pushes the model to generalize to these augmented query samples well after even one gradient update on $\mathcal{S}$, reducing the number of undesirable oscillations when predicting outside the samples from few referential compositions.
Formally, $\theta$ are updated as
\begin{align}
\label{eq7}  \theta^{'} = \theta - \epsilon \nabla_{\theta} \mathcal{L}_{\mathcal{S}}(f_\theta), \\
\label{eq8}  \theta \leftarrow \theta - \gamma \nabla_{\theta} \mathcal{L}_{\mathcal{\tilde{\mathcal{Q}}}}(f_{\theta^{'}}),
\end{align}
where hyper-parameters $\epsilon$ and $\gamma$ are the step size and the meta step size.
The experimental results in \tablename~\ref{tab:ablation} show that the combination of our Compositional Mixup and bi-level optimization can effectively handle the challenges of few-shot and few referential compositions, and thus improves the accuracy of recognizing unseen compositions.
We illustrate the pseudocode of the training and inference in Algorithm~\ref{algo:training_inference} for reference.
\vspace{-2mm}

\section{Experiments}    \label{experiments}

\begin{table*}[htbp]
  \centering
  \caption{Comparison with the baselines on two proposed benchmarks. Detailed results (\%) are reported with 95\% confidence intervals. Best results are displayed in \textbf{boldface}.}
  \vspace{-2mm}
  \resizebox{\linewidth}{!}{
    \begin{tabular}{c|ccccc|ccccc}
    \toprule
    \multirow{2}[2]{*}{\textbf{Method}} & \multicolumn{5}{c|}{\textbf{RL-CZSL-ATTR}} & \multicolumn{5}{c}{\textbf{RL-CZSL-ACT}} \\
          & UA & SA & HM & Prim 1 HM & Prim 2 HM & UA & SA & HM & Prim 1 HM & Prim 2 HM \\
    \midrule
    \multicolumn{1}{c}{} & \multicolumn{10}{c}{$K^c_s=1$} \\
    \midrule
    VisProd~\cite{Misra:red-wine-to-red-tomato} & 1.24 \mytiny{$\pm$0.43} & 20.99 \mytiny{$\pm$0.18} & 2.34 \mytiny{$\pm$0.76} & 25.44 \mytiny{$\pm$2.15} & 27.41 \mytiny{$\pm$1.79} & 0.88 \mytiny{$\pm$0.19} & 21.97 \mytiny{$\pm$0.56} & 1.68 \mytiny{$\pm$0.36} & 26.12 \mytiny{$\pm$1.78} & 25.43 \mytiny{$\pm$1.15} \\
    LE~\cite{Misra:red-wine-to-red-tomato} & 1.01 \mytiny{$\pm$0.96} & 14.98 \mytiny{$\pm$0.52} & 1.89 \mytiny{$\pm$1.67} & 22.25 \mytiny{$\pm$0.90} & 22.45 \mytiny{$\pm$0.53} & 1.27 \mytiny{$\pm$0.85} & 14.02 \mytiny{$\pm$0.18} & 2.32 \mytiny{$\pm$1.44} & 17.90 \mytiny{$\pm$3.28} & 22.45 \mytiny{$\pm$4.56} \\
    TMN~\cite{Purushwalkam:task-driven-modular-networks} & 0.52 \mytiny{$\pm$0.50} & \textbf{28.31} \mytiny{$\pm$0.39} & 1.02 \mytiny{$\pm$0.97} & 24.86 \mytiny{$\pm$1.59} & 32.41 \mytiny{$\pm$0.50} & 0.62 \mytiny{$\pm$0.34} & \textbf{28.85} \mytiny{$\pm$0.10} & 1.21 \mytiny{$\pm$0.65} & 31.76 \mytiny{$\pm$0.37} & 26.03 \mytiny{$\pm$1.55} \\
    SymNet~\cite{Li:symmetry-and-group} & 1.94 \mytiny{$\pm$0.08} & 17.34 \mytiny{$\pm$0.80} & 3.48 \mytiny{$\pm$0.12} & 27.01 \mytiny{$\pm$1.05} & 23.95 \mytiny{$\pm$2.87} & 2.28 \mytiny{$\pm$1.71} & 17.90 \mytiny{$\pm$0.56} & 4.01 \mytiny{$\pm$2.72} & 27.35 \mytiny{$\pm$1.59} & 23.02 \mytiny{$\pm$2.82} \\
    CompCos~\cite{Mancini:CompCos} & 2.57 \mytiny{$\pm$0.55} & 25.14 \mytiny{$\pm$0.70} & 4.66 \mytiny{$\pm$0.93} & 26.84 \mytiny{$\pm$0.72} & 33.53 \mytiny{$\pm$2.39} & 3.02 \mytiny{$\pm$0.34} & 28.19 \mytiny{$\pm$0.55} & 5.45 \mytiny{$\pm$0.56} & \textbf{32.07} \mytiny{$\pm$2.46} & \textbf{28.51} \mytiny{$\pm$2.57} \\
    CGE~\cite{Naeem:CGQA} & 4.65 \mytiny{$\pm$1.12} & 15.40 \mytiny{$\pm$0.54} & 7.13 \mytiny{$\pm$1.29} & 25.97 \mytiny{$\pm$3.30} & 31.56 \mytiny{$\pm$1.26} & 4.05 \mytiny{$\pm$0.78} & 15.51 \mytiny{$\pm$0.91} & 6.41 \mytiny{$\pm$0.91} & 28.56 \mytiny{$\pm$2.09} & 26.39 \mytiny{$\pm$1.50} \\
    \rowcolor[rgb]{ .949,  .949,  .949} \textbf{\shortOurs (Ours)} & \textbf{10.44} \mytiny{$\pm$0.42} & 19.01 \mytiny{$\pm$1.78} & \textbf{13.47} \mytiny{$\pm$0.77} & \textbf{30.22} \mytiny{$\pm$1.58} & \textbf{38.37} \mytiny{$\pm$2.29} & \textbf{7.76} \mytiny{$\pm$0.31} & 15.95 \mytiny{$\pm$1.24} & \textbf{10.44} \mytiny{$\pm$0.41} & 31.19 \mytiny{$\pm$1.38} & 26.68 \mytiny{$\pm$2.76} \\
    \midrule
    \multicolumn{1}{c}{} & \multicolumn{10}{c}{$K^c_s=5$} \\
    \midrule
    VisProd~\cite{Misra:red-wine-to-red-tomato} & 0.55 \mytiny{$\pm$0.45} & 15.83 \mytiny{$\pm$0.60} & 1.07 \mytiny{$\pm$0.83} & 22.80 \mytiny{$\pm$2.59} & 22.90 \mytiny{$\pm$1.31} & 0.18 \mytiny{$\pm$0.14} & 16.19 \mytiny{$\pm$0.32} & 0.35 \mytiny{$\pm$0.28} & 13.72 \mytiny{$\pm$1.80} & 21.88 \mytiny{$\pm$2.73} \\
    LE~\cite{Misra:red-wine-to-red-tomato} & 0.72 \mytiny{$\pm$0.49} & 13.79 \mytiny{$\pm$0.02} & 1.37 \mytiny{$\pm$0.89} & 21.14 \mytiny{$\pm$0.89} & 20.45 \mytiny{$\pm$2.64} & 1.23 \mytiny{$\pm$1.01} & 12.67 \mytiny{$\pm$0.62} & 2.23 \mytiny{$\pm$1.66} & 18.18 \mytiny{$\pm$2.41} & 18.88 \mytiny{$\pm$2.24} \\
    TMN~\cite{Purushwalkam:task-driven-modular-networks} & 0.27 \mytiny{$\pm$0.16} & \textbf{32.02} \mytiny{$\pm$0.51} & 0.54 \mytiny{$\pm$0.32} & 26.53 \mytiny{$\pm$0.60} & 35.22 \mytiny{$\pm$1.78} & 0.30 \mytiny{$\pm$0.29} & 31.28 \mytiny{$\pm$0.37} & 0.59 \mytiny{$\pm$0.57} & \textbf{34.03} \mytiny{$\pm$1.07} & 27.35 \mytiny{$\pm$1.43} \\
    SymNet~\cite{Li:symmetry-and-group} & 1.96 \mytiny{$\pm$0.95} & 18.47 \mytiny{$\pm$0.68} & 3.54 \mytiny{$\pm$1.54} & 27.24 \mytiny{$\pm$2.03} & 24.47 \mytiny{$\pm$2.57} & 2.28 \mytiny{$\pm$1.71} & 17.90 \mytiny{$\pm$0.56} & 4.01 \mytiny{$\pm$2.72} & 27.35 \mytiny{$\pm$1.59} & 23.02 \mytiny{$\pm$2.82} \\
    CompCos~\cite{Mancini:CompCos} & 1.05 \mytiny{$\pm$0.23} & 31.62 \mytiny{$\pm$0.54} & 2.03 \mytiny{$\pm$0.44} & 27.91 \mytiny{$\pm$2.68} & 34.71 \mytiny{$\pm$2.69} & 1.16 \mytiny{$\pm$0.55} & \textbf{34.32} \mytiny{$\pm$0.64} & 2.25 \mytiny{$\pm$1.02} & 32.59 \mytiny{$\pm$0.37} & 26.66 \mytiny{$\pm$2.82} \\
    CGE~\cite{Naeem:CGQA} & 4.10 \mytiny{$\pm$1.09} & 17.03 \mytiny{$\pm$0.13} & 6.61 \mytiny{$\pm$1.43} & 25.06 \mytiny{$\pm$0.20} & 31.04 \mytiny{$\pm$0.84} & 2.73 \mytiny{$\pm$0.78} & 19.12 \mytiny{$\pm$0.65} & 4.78 \mytiny{$\pm$1.17} & 25.57 \mytiny{$\pm$2.48} & 23.05 \mytiny{$\pm$1.11} \\
    \rowcolor[rgb]{ .949,  .949,  .949} \textbf{\shortOurs (Ours)} & \textbf{11.85} \mytiny{$\pm$2.55} & 20.70 \mytiny{$\pm$1.21} & \textbf{15.05} \mytiny{$\pm$1.81} & \textbf{31.88} \mytiny{$\pm$2.68} & \textbf{40.41} \mytiny{$\pm$1.25} & \textbf{8.01} \mytiny{$\pm$0.23} & 17.48 \mytiny{$\pm$0.98} & \textbf{10.99} \mytiny{$\pm$0.23} & 32.93 \mytiny{$\pm$1.73} & \textbf{28.01} \mytiny{$\pm$1.34} \\
    \bottomrule
    \end{tabular}%
    }
    \label{tab:all_results}%
    \vspace{-1mm}
\end{table*}%

\subsection{Experimental Setup}
For a fair comparison, the basic experiments are conducted with a four-layer convolution backbone (Conv-4) as in \cite{Chen:CloserLookFSC} for all implemented methods, and the backbone is fixed after training on the base data.
We also show the results on deeper backbones including ResNet-10 and ResNet-18~\cite{He:ResNet} in Section~\ref{sec:backbone} with the reason why we do not take them as the default choice.
If not specified, $K^c_s$, $K^c_q$, and $N^p$ are all set by default to 5 while $N^c_s$ and $N^c_q$ are dynamic and randomly sampled in each episode.   %
For methods using side information, we initialize the word embeddings with pre-trained 300-dimensional word2vec~\cite{Mikolov:word2vec} vectors.
And the best model is selected with the HM performance on the validation set.
The reported results are obtained by averaging 3 random experiments.
Code and datasets will be available at \url{https://github.com/bighuang624/RL-CZSL}.
\vspace{-2mm}

\paragraph{Pretraining Settings}
To pretrain the backbone, we use an Adam~\cite{Kingma:Adam} optimizer for the Conv-4 network and the stochastic gradient descent (SGD) optimizer for other backbone networks.
The backbone network, appended with a softmax layer, is trained with base data $\mathcal{D}_b$ to classify all compositions in $\mathcal{C}_b$ using the cross-entropy loss.
Standard data augmentation including random crop, left-right flip, and color jitter, is applied.
The pretraining lasts for a maximum of 500 epochs with a batch size of 128.
And the initial learning rate is set to $10^{-3}$ with a L2 penalty of $5 \times 10^{-4}$.
\vspace{-2mm}

\paragraph{Training Details}
Referring to~\cite{Chen:CloserLookFSC}, for methods that require training parameters in test episodes, we use the entire support set to train for 100 iterations with a batch size of 4.
All CZSL methods are trained with a SGD optimizer with an initial learning rate $10^{-2}$ and a L2 penalty of $10^{-3}$.
And we keep the other settings of hyperparameters in the public source code.
For our \shortOurs, we train at most 60,000 episodes with an initial learning rate $10^{-3}$ and a L2 penalty of $5 \times 10^{-4}$.
And the step size $\epsilon$ is set to 0.4.
Standard data augmentation is also applied when training all methods.
We implement our experiments in PyTorch~\cite{Paszke:PyTorch} and use a Nvidia V100 GPU to train all models.
\vspace{-2mm}

\paragraph{Unchosen Baselines}
We here discuss several existing CZSL methods that we did not include in our experiments. 
Attributes-as-operators (ATTOP)~\cite{Nagarajan:attributes-as-operators} views attributes as linear operators in the embedding space of object label embeddings.
As ATTOP is a method that often participates in comparison, we found it impossible to adapt to \shortname, and the main reason is the benchmark datasets, especially \datasetact, fail to provide antonyms for the method to calculate a loss term that operates over pairs of antonym attributes.
Besides, the commutative loss and the inverse loss cannot be calculated under the constraint of few referential compositions.
Some recent CZSL methods including OADis~\cite{Saini:OADis}, SCEN~\cite{Li:SCEN} and IVR~\cite{Tian:IVR} rely on simultaneously sampling images with same object and different attribute (or same attribute and different object) when updating the model, which is also unavailable in \shortname as one primitive may only appear in one composition in the episode.
Another compositional learning method proposed by \cite{Kato:HOI} was also considered at first, which is a rare HOI learning method that does not employ a pre-trained detector.
However, we found the authors did not release the source code, 
making us unable to implement.
\vspace{-2mm}

\subsection{Comparing with Baselines}
We compare \shortOurs and CZSL baseline methods on the two \shortname benchmark datasets.
As shown in \tablename~\ref{tab:all_results}, \shortOurs attains the best HA and UA. 
Although TMN and CompCos achieve better SA, the poor UA of all CZSL methods reveals that they overfit to seen compositions and fail to generalize to unseen ones when the number of referential compositions and samples is limited.

\begin{figure}[!t]    %
  \centering
  \includegraphics[width=0.48\textwidth]{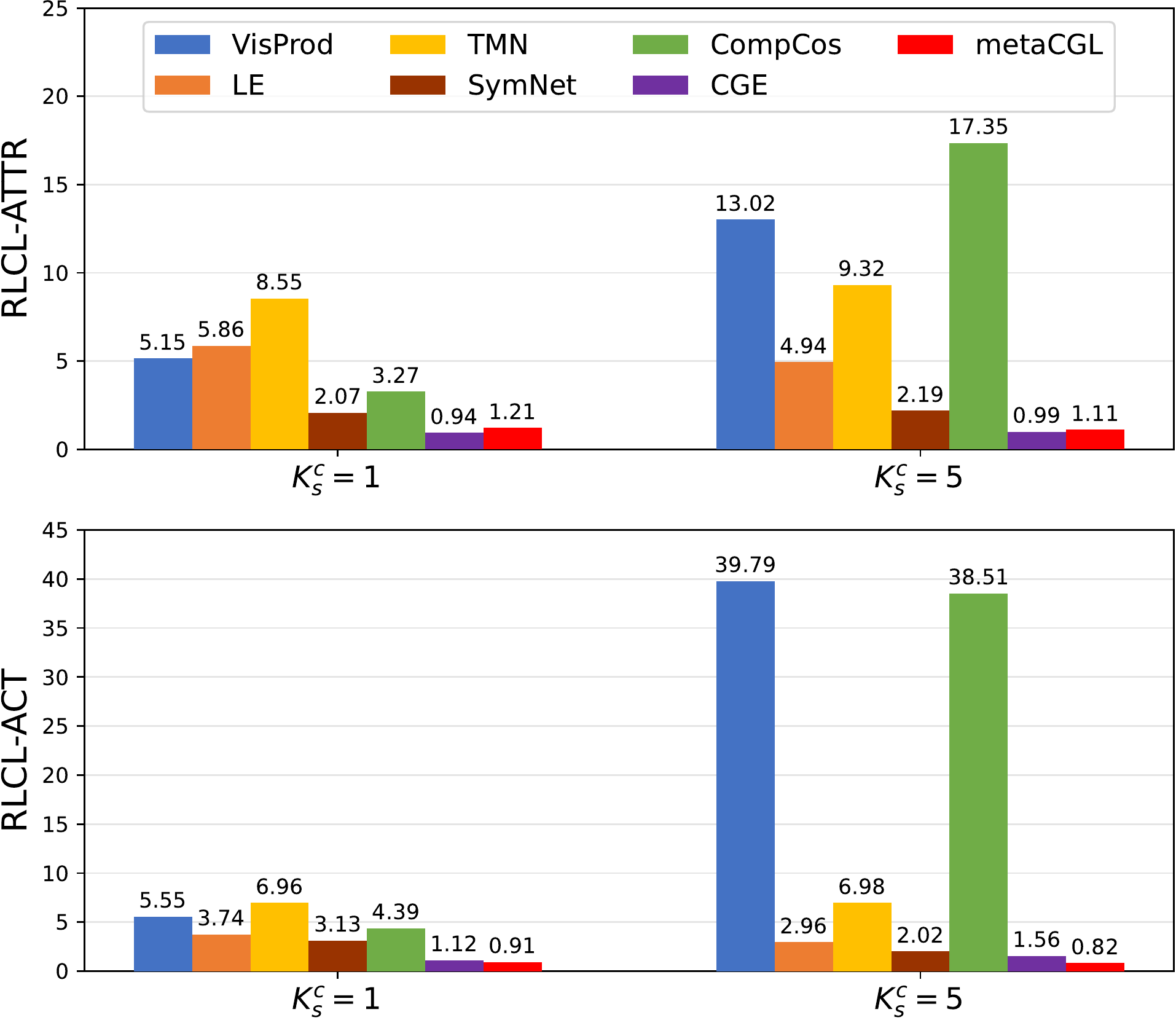}
  \vspace{-6mm}
  \caption{
    The ratios of whether error cases from unseen compositions are confused for seen pairs or incorrect unseen pairs. The lower the value, the smaller the trend of overfitting. Best viewed in color.
  }
  \label{fig:prediction_ratio}
\end{figure}

\subsection{Error Analysis}
To gain a further intuition of the performance of all methods, we analyze their errors when recognizing samples of unseen pairs.
Specifically, we calculate the metric of the prediction ratio of error cases from unseen compositions as $\frac{U \rightarrow S}{U \rightarrow U}$, where $S$ is the number of error cases that are confused for seen pairs, and $U$ represents the number of error cases that are confused for incorrect unseen pairs.
Different from the evaluation metrics used in the previously reported results, $\frac{U \rightarrow S}{U \rightarrow U}$ reflects the tendency of the model to predict seen compositions when encountering samples belonging to unseen compositions, \ie, it shows the degree of overfitting to seen compositions.
Therefore, although $\frac{U \rightarrow S}{U \rightarrow U}$ can not be directly used to represent the performance of the model, we expect that for a better compositional learner, this metric should be lower.
We illustrate the results in \figurename~\ref{fig:prediction_ratio}, and it can be observed that methods with better UA, like TMN and CompCos, also have a relatively high $\frac{U \rightarrow S}{U \rightarrow U}$ especially when $K^c_s$ increases. 
This indicates that they are poor at compositional learning, so increasing the sample size makes them tend to overfit the seen compositions.
And \shortOurs keeps a low $\frac{U \rightarrow S}{U \rightarrow U}$, especially on \datasetact.
Taken in conjunction with the results from \tablename~\ref{tab:all_results}, this suggests that our \shortOurs is a more positive compositional learner.
\vspace{-2mm}

\begin{table}[!t]  %
  \centering
  \caption{Ablation study with various model configurations of \shortOurs. $\mathcal{G}$: the compositional graph. $\mathcal{M}$: the correlation map generating network. BO: the bi-level optimization strategy. CM: the Compositional Mixup data augmentation.} %
  \vspace{-2mm}
  \resizebox{0.86\linewidth}{!}{
    \begin{tabular}{l|ccc|ccc}
    \toprule
    \multirow{2}[2]{*}{\textbf{Model}} & \multicolumn{3}{c|}{\textbf{RL-CZSL-ATTR}} & \multicolumn{3}{c}{\textbf{RL-CZSL-ACT}} \\
          & UA    & SA    & HM    & UA    & SA    & HM \\
    \midrule
    w/o $\mathcal{G}$ & 8.64  & 11.90 & 9.99  & 5.16  & 6.55  & 5.77 \\
    w/o $\mathcal{M}$ & 11.97 & 20.13 & 15.02 & 7.23  & 16.15 & 9.99 \\
    w/o BO & 1.25  & 13.70 & 2.28  & 0.85  & 13.88 & 1.60 \\
    w/o CM & 10.82 & \textbf{20.81} & 14.23 & 6.52  & \textbf{19.61} & 9.78 \\
    \rowcolor[rgb]{ .949,  .949,  .949} Full  & \textbf{11.85} & 20.70 & \textbf{15.05} & \textbf{8.01} & 17.48 & \textbf{10.99} \\
    \bottomrule
    \end{tabular}%
    }
  \label{tab:ablation}%
\end{table}%

\begin{table*}[!t]  %
  \centering
  \caption{Comparison with CZSL baselines equipped with MAML. Our \shortOurs still achieves the best UA and HM.}
  \vspace{-2mm}
  \resizebox{0.8\linewidth}{!}{
    \begin{tabular}{l|ccc|ccc|ccc|ccc}
    \toprule
    \multicolumn{1}{c|}{\multirow{3}[4]{*}{\textbf{Method}}} & \multicolumn{6}{c|}{\textbf{$K^c_s=1$}}   & \multicolumn{6}{c}{\textbf{$K^c_s=5$}} \\
\cmidrule{2-13}          & \multicolumn{3}{c|}{\textbf{RL-CZSL-ATTR}} & \multicolumn{3}{c|}{\textbf{RL-CZSL-ACT}} & \multicolumn{3}{c|}{\textbf{RL-CZSL-ATTR}} & \multicolumn{3}{c}{\textbf{RL-CZSL-ACT}} \\
          & UA    & SA    & HM    & UA    & SA    & HM    & UA    & SA    & HM    & UA    & SA    & HM \\
    \midrule
    VisProd~\cite{Misra:red-wine-to-red-tomato} & 1.24  & 20.99 & 2.34  & 0.88  & 21.97 & 1.68  & 0.55  & 15.83 & 1.07  & 0.18  & 16.19 & 0.35 \\
    \qquad+MAML & 3.53  & 20.76 & 6.03  & 1.72  & 21.50 & 3.18  & 3.76  & 29.71 & 6.67  & 2.23  & 28.60 & 4.13 \\
    \midrule
    LE~\cite{Misra:red-wine-to-red-tomato} & 1.01  & 14.98 & 1.89  & 1.27  & 14.02 & 2.32  & 0.72  & 13.79 & 1.37  & 1.23  & 12.67 & 2.23 \\
    \qquad+MAML & 4.49  & 6.24  & 5.21  & 6.14  & 7.93  & 6.91  & 6.06  & 14.15 & 8.48  & 5.68  & 8.06  & 6.64 \\
    \midrule
    SymNet~\cite{Li:symmetry-and-group} & 1.94  & 17.34 & 3.48  & 2.28  & 17.90 & 4.01  & 1.96  & 18.47 & 3.54  & 2.96  & 17.12 & 5.04 \\
    \qquad+MAML & 3.62  & 4.48  & 4.00  & 4.91  & 4.47  & 4.65  & 3.40  & 4.14  & 3.69  & 3.61  & 4.74  & 4.10 \\
    \midrule
    CompCos~\cite{Mancini:CompCos} & 2.57  & \textbf{25.14} & 4.66  & 3.02  & \textbf{28.19} & 5.45  & 1.05  & \textbf{31.62} & 2.03  & 1.16  & \textbf{34.32} & 2.25 \\
    \qquad+MAML & 3.17  & 5.93  & 4.07  & 2.81  & 6.44  & 3.90  & 2.98  & 7.67  & 4.28  & 3.54  & 8.76  & 5.02 \\
    \midrule
    CGE~\cite{Naeem:CGQA} & 4.65  & 15.40 & 7.13  & 4.05  & 15.51 & 6.41  & 4.10  & 17.03 & 6.61  & 2.73  & 19.12 & 4.78 \\
    \qquad+MAML & 9.44  & 18.62 & 12.54 & 6.06  & 17.86 & 9.05  & 11.09 & 19.76 & 14.21 & 6.73  & 18.17 & 9.82 \\
    \midrule
    \rowcolor[rgb]{ .949,  .949,  .949} \textbf{\shortOurs (Ours)} & \textbf{10.44} & 19.01 & \textbf{13.47} & \textbf{7.76} & 15.95 & \textbf{10.44} & \textbf{11.85} & 20.70 & \textbf{15.05} & \textbf{8.01} & 17.48 & \textbf{10.99} \\
    \bottomrule
    \end{tabular}%
    }
    \label{tab:CZSL_with_MAML}
    \vspace{-1mm}
\end{table*}%

\subsection{Ablation Study} %
In \tablename~\ref{tab:ablation}, we examine the effectiveness of each component in \shortOurs.
The first observation is that the removal of any component from \shortOurs generally results in a worse performance on UA and HM, which upholds the efficacy of our framework design.
Among all the components, the compositional graph plays an important role in recognizing samples of seen compositions, and the bi-level optimization significantly contributes to recognizing unseen compositions.
Another phenomenon is that our Compositional Mixup method improves UA while sacrificing SA, and it is understandable as the data augmentation inhibits the further use of samples from seen compositions to evaluate the quality of the fast adaptation, so as to combat memorization of seen compositional labels and alleviate the overfitting to seen compositions.
\vspace{-2mm}

\subsection{Effect of Backbone Network} \label{sec:backbone}

\begin{figure}[!t]    %
  \centering
  \includegraphics[width=0.46\textwidth]{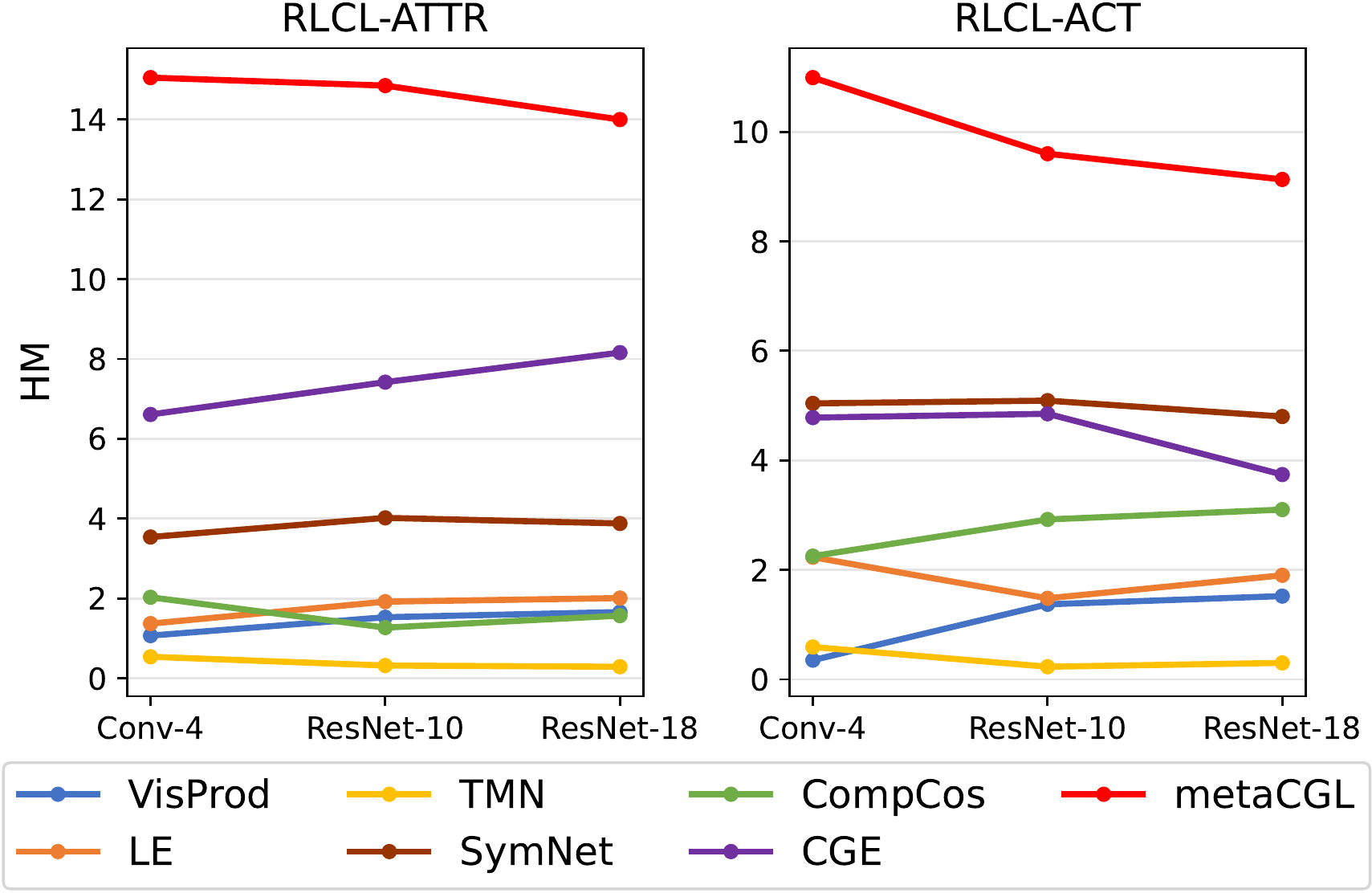}
  \vspace{-2mm}
  \caption{
    Ablation results of different backbone networks. Deeper backbone does not necessarily lead to better performance. Best viewed in color. 
  }
  \label{fig:backbone}
\end{figure}

Comparing to our default backbone Conv-4 that is often used in FSL, existing CZSL works prefer to use a deeper ResNet~\cite{He:ResNet} backbone network, \eg, ResNet-18 in \cite{Naeem:CGQA}.
Moreover, \cite{Chen:CloserLookFSC} proposes that increasing the depth of the backbone network improves the FSL methods by reducing intra-class variation.
Therefore, we conduct an ablation study by increasing the backbone network from Conv-4 to ResNet-10 and ResNet-18 when $K^c_s=5$, exploring whether this can also improve the performance in \shortname.
Specifically, ResNet-18 is the same as described by \cite{He:ResNet} with an input size of $84\times84$, while ResNet-10 is a simplified version where only one residual building block is used in each layer.
We illustrate the results in \figurename~\ref{fig:backbone}, and it can be observed that while the tendency of the same method is quite unstable on different datasets, different methods also show no consistent pattern on the same dataset when the backbone deepens. 
In other words, a deeper backbone network will not necessarily result in a boost in \shortname if not paired with the suitable approach.
\vspace{-2mm}

\subsection{Effect of Equipping MAML}

From the ablation study results of our \shortOurs, it can be observed that the bi-level optimization significantly contributes to recognizing unseen compositions.
Therefore, to study whether such bi-level optimization can also improve the baselines on the \shortname task, we equip these methods with the popular bi-level optimization method named model-agnostic meta-learning (MAML)~\cite{Finn:MAML}.
Note that TMN~\cite{Purushwalkam:task-driven-modular-networks} is not considered as the bi-level optimization will double its large computation overhead, making the computing time unaffordable.
We compare the results with our \shortOurs in \tablename~\ref{tab:CZSL_with_MAML}.
It can be observed that, on the one hand, VisProd and CGE consistently gain from equipping MAML. 
On the other hand, MAML significantly hurts the seen accuracy of other methods (LE, SymNet, and CompCos) without bringing a stable gain in unseen accuracy, and the performance of SymNet even degenerates to close to that of random prediction.
On top of that, our \shortOurs still achieves the best UA and HM in all cases.
The results of this experiment show that the bi-level optimization is not a silver bullet for the \shortname problem, and more effective solutions remain to be explored.
\vspace{-2mm}

\section{Conclusion}

In this paper, we introduce \setting (\shortname), a novel and non-trivial task that mimics the naturalistic unseen environment for compositional learners.
Aiming to recognize unseen compositional concepts in the scarcity of referential compositions and samples, we propose a \Ours (\shortOurs) that can efficiently learn the compositionality in the new environment.
Moreover, we also build two large-scale benchmark datasets to drive research on the task.
We show that in the challenging \shortname setting, our \shortOurs significantly outperforms the state-of-the-art CZSL methods in recognizing unseen compositions, while substantial research space still remains.
We hope our work can facilitate and calibrate the development of compositional learning systems. %

\noindent \textbf{Acknowledgement}
This work was supported by STI 2030—Major Projects (2022ZD0208800), and NSFC General Program (Grant No. 62176215).

\bibliographystyle{ACM-Reference-Format}
\bibliography{sample-base}

\end{document}